\documentclass[letterpaper]{article} 
\usepackage[draft]{aaai2026}
\usepackage{times}  
\usepackage{helvet}  
\usepackage{courier}  
\usepackage[hyphens]{url}  
\usepackage{graphicx} 
\usepackage{natbib}  
\usepackage{caption} 
\usepackage{algorithm}
\usepackage{algorithmic}

\usepackage{arydshln}
\usepackage{multirow}
\usepackage{multicol}
\usepackage{graphicx}
\usepackage{subfig}
\usepackage{enumitem}
\usepackage{booktabs}
\usepackage{multirow}
\usepackage{newfloat}
\usepackage{listings}
\DeclareCaptionStyle{ruled}{labelfont=normalfont,labelsep=colon,strut=off} 
\floatstyle{ruled}
\newfloat{listing}{tb}{lst}{}
\floatname{listing}{Listing}
\title{Affordance Benchmark for MLLMs}
\author{Junying Wang$^{1,2}$, Wenzhe Li$^{2}$, Yalun Wu$^{3}$, Yingji Liang$^{2}$, Yijin Guo$^{2,3}$, \\
\textbf{Chunyi Li$^{2,3}$, Haodong Duan$^{2}$, Zicheng Zhang$^{2,\dagger}$, Guangtao Zhai$^{2,3,\dagger}$}\\
}
\affiliations{
$^1$Fudan University, $^2$Shanghai Artificial Intelligence Laboratory, $^3$Shanghai Jiao Tong University\\
 $^\dagger$Corresponding author. \\
{ \textit{Project Page}: https://github.com/JunyingWang959/A4Bench \enspace  \textit{Team}: https://aiben.ch}
 }
\usepackage{bibentry}

\begin{document}

\maketitle

\begin{figure*}[t]
  \centering
  \includegraphics[width=\linewidth]{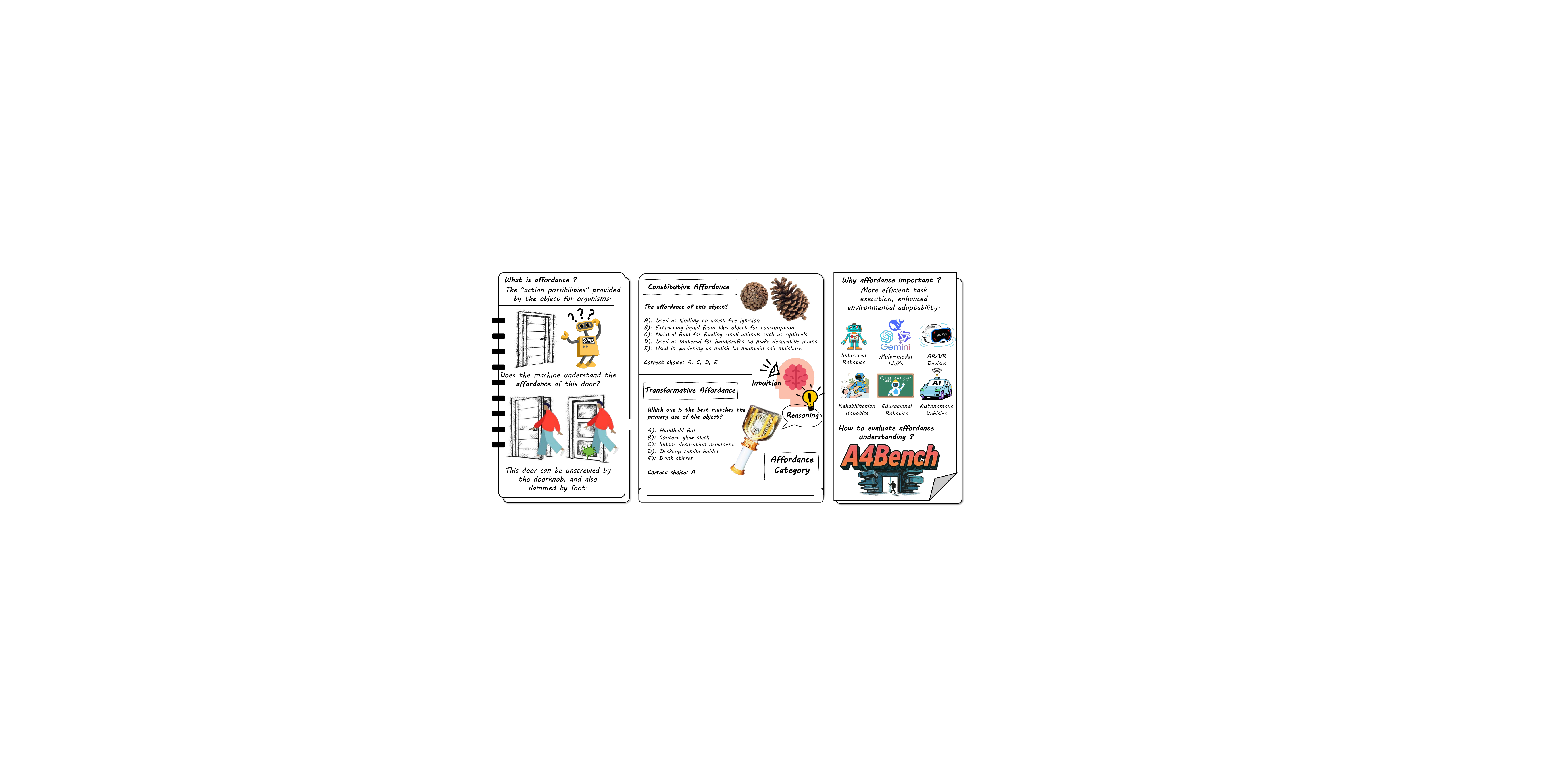}
  \caption{The motivation of the A4Bench. The affordance theory proposed by James J. Gibson\cite{gibson2014ecological} defines the action possibilities provided by the object for organisms. Evaluating the affordance perception abilities of MLLMs can help enable more efficient task execution and improved adaptability to diverse environments for AI systems.}
  \label{fig:spot}
\end{figure*}

\begin{abstract}
Affordance theory suggests that environments inherently provide action possibilities shaping perception and behavior. While Multimodal Large Language Models (MLLMs) achieve strong performance in vision-language tasks, their ability to perceive affordance, which is crucial for intuitive and safe interactions, remains underexplored. To address this, we introduce \textbf{A4Bench}, a novel benchmark designed to evaluate the affordance perception abilities of MLLMs across two dimensions: 1) \textit{Constitutive Affordance}, assessing understanding of inherent object properties through 1,282 question-answer pairs spanning nine sub-disciplines, and 2) \textit{Transformative Affordance}, probing dynamic and contextual nuances (e.g., misleading, time-dependent, cultural, or individual-specific affordance) with 718 challenging question-answer pairs. We evaluate 17 MLLMs (nine proprietary and eight open-source) and compare them to human performance. Results show that proprietary models generally outperform open-source ones, yet all models perform far below humans, especially in transformative affordance. Furthermore, even top-performing models, such as Gemini-2.0-Pro (18.05\% overall exact match accuracy), significantly lag behind human performance (best: 85.34\%, worst: 81.25\%). These findings highlight critical gaps in environmental understanding of MLLMs and provide a foundation for advancing AI systems toward more robust, context-aware interactions. 
\end{abstract}

\section{Introduction}

`\textit{What we perceive when we look at objects are their affordances, not their qualities}'. This bold assertion by James J. Gibson encapsulates the essence of \textit{affordance theory}~\cite{gibson2014ecological}, which argues that environments inherently provide organisms with a spectrum of action possibilities shaping perception and behavior. Unlike traditional theories that focus on physical attributes, affordance theory emphasizes the direct perception of \textit{action possibilities} embedded in environmental features.  For example, a door is not merely a static structure but something that affords multiple actions: it can be unlocked with a key, pushed open, or even kicked. As illustrated in Figure~\ref{fig:spot} (bottom left), affordances go beyond conventional functionality to encompass a wide range of potential interactions grounded in the material, shape, and context of an object. 
Thus, affordances form a crucial link between ecological features and behavioral responses, supporting a unified ecological approach to perception.

\textbf{\textit{Why affordance is important?}}

Understanding affordance is vital for intelligent agents, whether biological or artificial, to \textbf{engage meaningfully with their environments}. For artificial intelligence systems navigating complex settings, understanding affordance ensures robust and intuitive interactions while enhancing safety. For example, a robot perceiving a surface as affording support can navigate terrain securely, while one identifying an object as graspable can manipulate it effectively. Moreover, effective affordance perception enables industrial robots to \textbf{execute tasks with greater efficiency} and allows rehabilitation robots to \textbf{enhance human-machine interaction} by adapting to user needs. The capacity to discern affordance, whether beneficial or harmful, underpins behaviors ranging from survival to complex social interactions. Gibson underscored this by noting how humans modify environments 'to change what it affords', emphasizing the profound link between affordance and intentional action.


\textbf{\textit{What Affordance Perception Should MLLMs Possess?}}

Accurate perception of an object affordance enables recognition of its true utility, as Gibson asserts, `{If the affordance of a thing are perceived correctly, we say that it looks like what it is}'. Yet discerning genuine affordance often demands experiential learning, since a leaf appearing benign may conceal a nettle sting or a politician seeming helpful may mask deceptive demagoguery.

To evaluate this capability rigorously, we introduce a novel benchmark (\textbf{A4Bench}) to assess affordance perception across two primary dimensions. First, \textbf{constitutive affordance} examines how MLLMs apprehend inherent object and environmental properties such as shape, size, or material that determine whether a surface affords walking or an object affords grasping, using 1282 question-answer pairs across nine sub-disciplines. Second, \textbf{transformative affordance} probes comprehension of dynamic affordance, including \textbf{misleading affordance} (like a glass barrier appearing as open air but affording collision), \textbf{time-independent affordance} (e.g., fruit ripening changes its food value affordance, specified by color), \textbf{cross-cluture affordance} (e.g., a postbox affording letter-mailing only within a postal system), and \textbf{individual affordance} (e.g., a ledge being sit-on-able depending on leg length). 

\textbf{A4Bench} pioneers comprehensive affordance perception evaluation, highlighting contextual and dynamic nuances to reveal MLLM capabilities and steer future development. Testing 17 leading MLLMs, both open-source and closed-source, on \textbf{A4Bench} reveals that even the top-performing model significantly trails human performance, leading to a critical conclusion:

\textbf{\textit{MLLMs are still poor at affordance perception.}}

MLLMs exhibit significant limitations in perceiving affordance with human-like proficiency, with marked disparities between open-source and closed-source variants. This deficiency spans constitutive and transformative affordance. Challenges arise from limited contextual understanding, notably in medical disciplines within constitutive affordance and agent-specific dimensions within transformative affordance. Conversely, human observers consistently excel across all dimensions, highlighting inadequate comprehension by MLLMs of object-human-environment interactions essential for robust affordance perception (poor at affordance perception).


Evaluating affordance perception in MLLMs shapes AI advancements by driving \textbf{A4Bench} development, as shown in Figure \ref{fig:spot}. And our contributions are summarized as:

\begin{itemize}[leftmargin=1.5em]
    \item We extend affordance theory to the context of MLLMs and introduce a systematic evaluation framework for affordance that rigorously tests the understanding of MLLMs on understanding constitutive and transformative affordance. 
    \item We propose \textbf{A4Bench}, the first comprehensive benchmark specifically designed to assess the affordance perception capabilities of MLLMs. It encompasses 2,000 multimodal question-answer pairs, covering both Constitutive Affordance (1,282 pairs across nine sub-disciplines) and Transformative Affordance (718 pairs addressing misleading, time-dependent, cross-cultural, and individual-specific affordance).
    \item By evaluating 17 MLLMs (9 proprietary and 8 open-source) against human performance, we provide a detailed analysis of their affordance perception capabilities. Our findings reveal significant limitations in existing models, with even top performers lagging far behind human benchmarks, thus offering a roadmap for future improvements in context-aware AI systems.
\end{itemize}

\begin{figure*}[t]\small
    \centering
    \begin{minipage}{0.62\textwidth}
        \includegraphics[width=\linewidth]{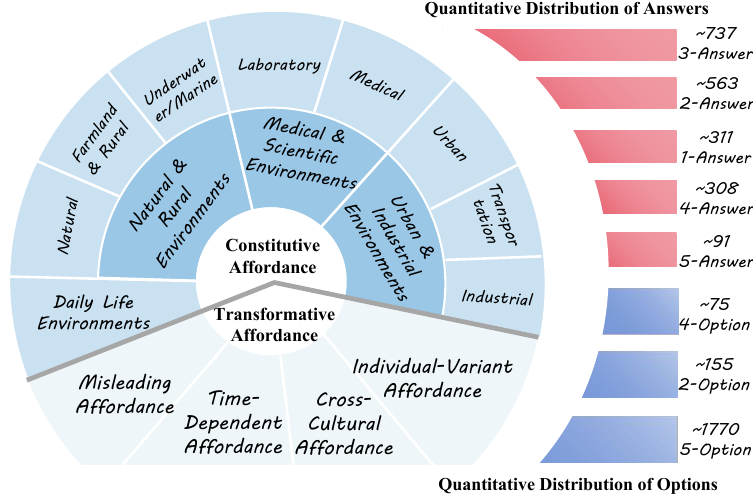}
    \end{minipage}\hfill
    \begin{minipage}{0.34\textwidth}
        \centering
        \begin{tabular}{lc}
            \toprule
            \renewcommand\arraystretch{1.2}
            \bf A4Bench    &\textbf{2,000}\\
            \midrule
            \textbf{Constitutive Affordance} & \textbf{1,282}   \\
            $\quad$Natural \& Rural Environments & 423 \\
            $\quad$$\quad$Natural  & 220 \\
            $\quad$$\quad$Farmland \& Rural & 81 \\
            $\quad$$\quad$underwater / Marine & 116 \\
            $\quad$Medical \& Scientific Environments & 245 \\
            $\quad$$\quad$Laboratory  & 119 \\
            $\quad$$\quad$Medical  & 126 \\
            $\quad$Urban \& Industrial Environments & 388 \\
            $\quad$$\quad$Urban  & 170 \\
            $\quad$$\quad$Transportation & 118 \\
            $\quad$$\quad$Industrial & 100 \\
            \midrule
            \textbf{Transformative Affordance} & \textbf{718}   \\
            $\quad$Misleading Affordance & 187 \\
            $\quad$Time-Dependent Affordance & 210 \\
            $\quad$Cross-Cultural Affordance & 118 \\
            $\quad$Individual-Variant Affordance & 203 \\
            \bottomrule
        \end{tabular}
    \end{minipage}
    \caption{Structure and quantitative overview of \textbf{A4Bench}. The left panel presents the focused aspects of the benchmark, detailing the primary dimensions (Constitutive and Transformative Affordance) and their respective sub-dimensions. The middle panel depicts the distributions of answer counts and option counts. The right panel reports the number of multimodal question–answer pairs across each sub-dimension, providing a comprehensive overview of the dataset composition.}
    \label{fig:dis}
\end{figure*}

\section{Related Works}
\paragraph{\textbf{Multimodal Large Language Models.}}
Multimodal Large Language Models (MLLMs) have demonstrated remarkable capabilities in diverse vision-language tasks. Among them, diverse proprietary (\textit{close-source}) models \citep{openai2024gpt4o, anthropic2024claude3,google2024gemini,Step-1o,BailingMM-Pro-0120,MUG-U-7B}  and representative open-source models 
\citep{wu2024janus,team2025kimi,dai2023instructblip,zhu2023minigpt4,liu2023visual,liu2024improved,li2024llava,lu2024ovis,guo2025seed1,dong2025scalable,yao2024minicpm,dai2024nvlm,agrawal2024pixtral,liu2024points,zhang2025flash,lin2023vila,chen2025janus} have exhibited impressive superiority from embodied agents \citep{song2023llmplanner} to other real-world applications \citep{huynh2025visual, 10684794, jiao2025cnn2gnn,10412657}. However, whether these
MLLMs are \textbf{masters at percepting the affordance of object} is still questionable, which needs further investigation.


\paragraph{\textbf{Multimodal Benchmarks.}}
Current benchmarks provide valuable insights, assessing capabilities ranging from bilingual perception and reasoning \cite{liu2025mmbench}, hierarchical comprehension\cite{li2024seed,EESE,interview,redundancy_principle}, expert-level multimodal tasks \cite{2024mmt-icml,10870403,hce,qalign,qbench,9810024}, and complex embodied scenarios \cite{chen-etal-2024-pca}, to critical nuances like visual dependency \cite{NEURIPS2024}, safety against adversarial inputs \cite{2024-safety-eccv, Wang_2025_CVPR, aibench}, and understanding of AI-generated images \cite{zhang2024a-bench}. Despite these efforts, there is still a gap in explicitly assessing the affordance perception abilities of MLLMs. Since understanding affordance is fundamental for any intelligent agent to interact effectively, meaningfully, and safely with its surroundings, \textbf{{A4Bench}} is developed to bridge this gap.


\section{Constructing the A4Bench}
\subsection{Key Principles}

\paragraph{\textbf{Covering Constitutive and Transformative Affordance}}
\textit{The human brain, when visually recognizing action interactions, processes static forms alongside dynamic movements} \cite{blake2007perception}. Inspired by this, assessing MLLMs in perceiving object affordance involves \textit{understanding inherent object properties and their empowering potential}. Assessing whether these models meet such criteria requires examining proficiency in (\textbf{constitutive affordance}) and (\textbf{transformative affordance}) perception. As shown in Figure \ref{fig:dis}, constitutive affordance covers a diverse range of scenarios, while transformative affordance includes misleading affordance, individual-variant affordance, time-dependent affordance, and cross-cultural affordance.


\paragraph{\textbf{Guaranteeing Benchmarking Difficulty} }
\textit{Rigorous benchmarking effectively evaluates rapid advancements in MLLMs at the forefront of human knowledge} \cite{phan2025humanitysexam}. To ensure the difficulty and quzlity of benchmark, diverse strategies are implemented: a quality control mechanism, a vision-language prompt removal approach, and an unspecified options and answers strategy. 
1) \textbf{Quality control mechanism} employs a human-MLLM mixed-obfuscation adaptation process. This iterative method starts with human-generated problems, followed by alternating revisions between models and experts until the model fails to respond correctly, ensuring a challenging benchmark.
2) \textbf{Vision-language prompt removal approach} enhances affordance comprehension assessment by replacing explicit object names in images, questions, and options with `this object.' This method replicates real-world multimodal perception environments critical for affordance evaluation while increasing difficulty for robust model assessment.
3) \textbf{Unspecified options and answers strategy} heightens challenge by concealing the exact number of correct answers from models and participants, thus reducing reliance on guessing while prioritizing deep comprehension. And the quantitative distributions of the answers and options are shown in the middle panel of Figure \ref{fig:dis}.


\begin{figure*}[h]
  \centering
  \includegraphics[width=\linewidth]{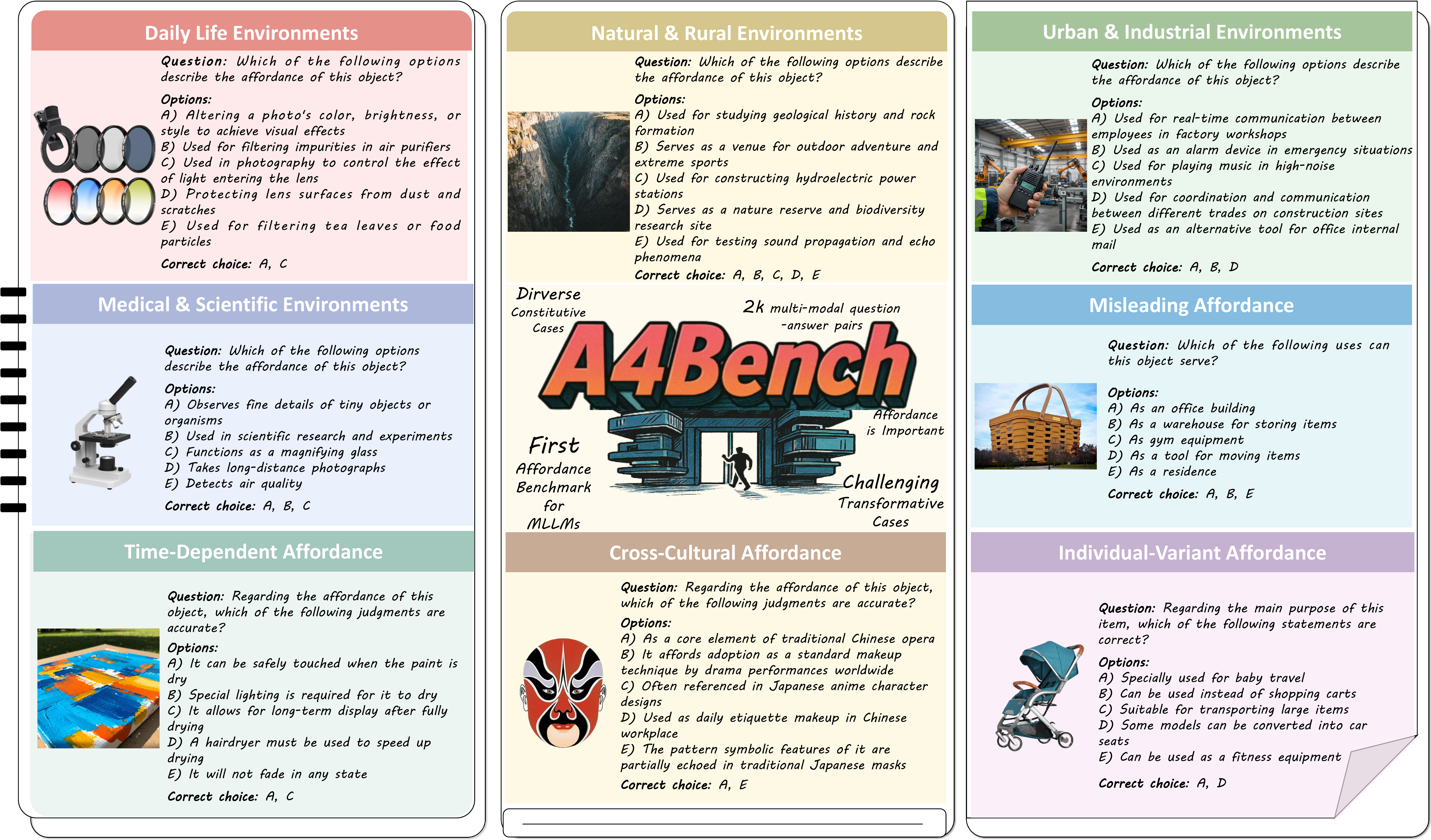}
  \caption{ Typical samples from the \textbf{A4Bench}. Each sample is accompanied by a image-question-answer pair. \textbf{A4Bench} evaluates models across
diverse discplines (Constitutive Affordance) and challenging dimensions (Transformative Affordance), ensuring a comprehensive evaluation of the affordance perception capabilities.}
  \label{fig:case}
\end{figure*}

\subsection{Focused Aspects}
Evaluating affordance perception examines inherent properties such as structure and material alongside potential dynamic action interactions, encompassing constitutive and transformative affordance dimensions. And the representative examples of these multifaceted dimensions are clearly and visually illustrated in Figure \ref{fig:case}.

\paragraph{\textbf{Constitutive Affordance Perception}}
We propose \textit{A4Bench}, a robust benchmark designed to assess affordance perception in multimodal large language models. To evaluate basic affordance comprehension, the \textbf{Constitutive Affordance} component includes 1282 multimodal question-answer pairs across four disciplines and nine sub-disciplines, emphasizing inherent object properties such as shape, size, and material that determine walkability or graspability. These pairs probe static environmental cues critical for accurate affordance detection. The disciplines \textbf{Daily Life Environments} (226 pairs), \textbf{Natural and Rural Environments} (423 pairs), \textbf{Urban and Industrial Environments} (388 pairs), and \textbf{Medical and Scientific Environments} (245 pairs) comprise 64.1\% of the benchmark. And the details of sub-discipline are illustrated in Figure \ref{fig:dis}.

\paragraph{\textbf{Transformative Affordance Perception.}}
To evaluate comprehension of dynamic object potential in MLLMs, the \textbf{Transformative Affordance} component of \textit{A4Bench} employs 718 challenging multimodal question-answer pairs designed to probe complex context-sensitive affordance perception across diverse real-world scenarios. This component rigorously tests the ability of MLLMs to interpret nuanced interactions beyond static properties, ensuring robust evaluation of adaptive affordance understanding critical for practical applications. These pairs, categorized into four distinct aspects, address high-difficulty dynamic interactions: 
1) \textbf{Misleading Affordance Perception} examines visually deceptive objects such as pillows resembling fire hydrants or phone cases mimicking slippers, testing perceptual accuracy under visual ambiguity in everyday contexts.
2) \textbf{Time-Dependent Affordance Perception} investigates affordance evolving over time, exemplified by a concrete surface transitioning from impassable to walkable after six months of curing or fruit ripening to edibility, reflecting temporal dynamics in natural processes.
3) \textbf{Cross-Cultural Affordance Perception} analyzes variations due to cultural contexts, where a thumbs-up gesture conveys approval in some cultures but signals taxi-hailing in others, highlighting cultural specificity in global interactions.
4) \textbf{Individual-Variant Affordance Perception} explores affordance differing by individual, such as a stroller providing seating for infants but not adults, emphasizing tailored interaction potential across populations with distinct needs.

\begin{table*}[t]\small
    \centering
    \renewcommand\arraystretch{1.1}
    \setlength{\tabcolsep}{2pt}
    \caption{{ Benchmark results on the \textbf{A4Bench}, displaying \textit{Exact match Accuracy}, with the best performance marked in \textbf{bold} and the second-best \underline{underlined} for both proprietary and open-source MLLMs. The \textit{Overall score of Constitutive Affordance} is the weighted average of Daily, Natural, Medical, and Urban discplines. The \textit{Overall score of Transformative Affordance} averages Time-Indenpendent, Cross-Culture, and Individual dimensions. The \textit{final Overall score} encompasses all pairs.}}
    
    \begin{tabular}{l|ccccc|c|cccc|c}
    \hline
    \hline
         \textbf{Categories} & \multicolumn{5}{c|}{\textbf{Constitutive Affordance}} & \multicolumn{5}{c|}{\textbf{Transformative Affordance}}  &{\textbf{Overall$\uparrow$}}\\ \cdashline{1-12}
        {\textbf{MLLM} \textit{(MLLM)}}  & {\textit{Daily$\uparrow$}}& {\textit{Natural$\uparrow$}} & {\textit{Medical$\uparrow$}} & {\textit{Urban$\uparrow$}} & {\textit{Overall$\uparrow$}} & \textit{Misleading$\uparrow$}  & \textit{Time$\uparrow$}  & \textit{Culture$\uparrow$} & \textit{Individual$\uparrow$} & {\textit{Overall$\uparrow$}} \\\hline
        \textsc{Human (Best)}  & 83.63\% & 87.35\% & 83.39\% & 87.50\% & 85.99\% & 87.17\% & 83.67\% & 84.75\% & 81.66\% & 83.14\% & 85.34\% \\
        \textsc{Human (Worst)} & 79.20\% & 85.71\% & 78.83\% & 81.55\% & 81.88\% & 83.96\% & 81.43\% & 78.81\% & 76.03\% & 78.78\% & 81.25\% \\
       \hline
       \multicolumn{10}{l}{\textbf{Proprietary MLLMs:}} \\ \hdashline
    \textsc{o3}  & 14.64\% & 11.45\% & 10.28\% & 11.37\% & 11.76\% & 63.64\% & 10.95\% & 10.17\% & \underline{7.39\%} & 9.42\% & 15.99\% \\
    \textsc{ChatGPT-4o} & 16.41\% & 10.45\% & 9.63\% & 10.98\% & 11.52\% & 67.38\% & \textbf{13.33\%} & \underline{16.10\%} & \underline{7.39\%} & \underline{11.68\%} & 16.78\% \\
    \textsc{GPT-4.1} & 18.18\% & 10.45\% & 9.95\% & 9.99\% & 11.57\% & \underline{68.98\%} & 8.10\% & 10.17\% & 2.96\% & 6.59\% & 15.61\% \\
    \textsc{GPT-4.1-Mini} & 12.87\% & 10.86\% & 9.45\% & 9.79\% & 10.59\% & 63.10\% & 7.24\% & 14.58\% & 2.81\% & 7.18\% & 14.59\% \\
    \textsc{GPT-4o} & \textbf{19.95\%} & 10.45\% & 8.98\% & 9.39\% & 11.49\% & 65.24\% & 9.52\% & 9.32\% & 4.93\% & 7.72\% & 15.52\% \\
    \textsc{Claude-3.5-Sonnet} & 14.19\% & \underline{13.71\%} & 9.63\% & 10.38\% & 11.92\% & 61.50\% & 7.62\% & 14.41\% & 6.40\% & 8.66\% & 15.69\% \\
    \textsc{Claude-3.7-Sonnet} & 17.29\% & \textbf{17.80\%} & \underline{11.26\%} & \underline{12.96\%} & \textbf{14.86\%} & 45.99\% & 10.48\% & \textbf{24.58\%} & \textbf{12.32\%} & \textbf{14.31\%} & \underline{17.63\%} \\
    \textsc{Gemini-2.0-Flash} & 17.29\% & 10.04\% & 9.95\% & 9.59\% & 11.15\% & 68.45\% & 5.71\% & 11.02\% & 5.91\% & 6.97\% & 15.40\% \\
    \textsc{Gemini-2.0-Pro} & \underline{18.26\%} & \underline{13.71\%} & \textbf{14.19\%} & \textbf{14.35\%} & \underline{14.82\%} & 67.38\% & 9.52\% & 13.56\% & 4.43\% & 8.47\% & \textbf{18.05\%} \\
    \hline
    \multicolumn{10}{l}{\textbf{Open-source MLLMs:}} \\ \hdashline
    \textsc{DeepSeek-VL2} & 9.33\% & 9.63\% & 8.65\% & 8.79\% & 9.11\% & \textbf{69.50\%} & 8.10\% & 12.71\% & 3.97\% & 7.54\% & 14.34\% \\ 
    \textsc{DeepSeek-VL2-Small} & 9.77\% & 8.82\% & 8.65\% & 8.79\% & 8.95\% & 65.78\% & 6.19\% & 11.86\% & 2.46\% & 6.03\% & 13.48\% \\
    \textsc{DeepSeek-VL2-Tiny} & 9.77\% & 8.82\% & 8.65\% & 8.79\% & 8.95\% & 65.78\% & 6.13\% & 11.80\% & 2.43\% & 5.98\% & 13.47\% \\
    \textsc{InternVL3-14B} & 8.44\% & 8.41\% & 9.30\% & 9.19\% & 8.84\% & 60.43\% & \underline{11.43\%} & 15.25\% & 2.96\% & 9.04\% & 13.72\% \\
    \textsc{InternVL3-38B} & 9.33\% & 9.63\% & 9.30\% & 8.60\% & 9.17\% & 62.03\% & 7.62\% & 13.56\% & 2.96\% & 7.16\% & 13.58\% \\
    \textsc{MpLUG-OWL3-7B} & 10.21\% & 9.63\% & 8.33\% & 9.79\% & 9.54\% & 57.75\% & 7.62\% & 11.02\% & 0.99\% & 5.84\% & 13.06\% \\
    \textsc{Qwen2.5-VL-32B}& 8.88\% & 10.04\% & 9.30\% & 8.40\% & 9.15\% & 66.31\% & 7.14\% & 8.47\% & 1.97\% & 5.46\% & 13.52\% \\
    \textsc{Qwen2.5-VL-72B } & 10.65\% & 13.31\% & 9.95\% & 9.98\% & 11.10\% & 64.71\% & 8.10\% & 9.32\% & 5.42\% & 7.34\% & 15.12\% \\
        \hline         
        \textit{Random guess} & 1.33\% & 1.55\% & 1.22\% & 2.60\% & 1.79\% & 44.92\% & 1.42\% & 0.85\% & 0.05\% & 0.77\% & 5.55\% \\
        \hline
        \hline
    \end{tabular}
    
    \label{tab:part1}
\end{table*}


\subsection{Question Collection}
\paragraph{\textbf{Question Type}} 

In the \textbf{A4Bench}, two distinct question formats are employed: \textit{Yes-or-No} and \textit{What} questions. \textit{Yes-or-No} questions (7.8\%) evaluate fundamental judgment capabilities in MLLMs, while \textit{What} questions (92.2\%) demand deeper affordance comprehension due to their inherent complexity, facilitating a comprehensive and robust assessment of nuanced conceptual understanding.

\paragraph{\textbf{Human Expert Annotation}}

A team of 60 human annotators, categorized by professional expertise into 10 senior experts, 20 junior researchers, and 30 general researchers, develops questions for \textbf{A4Bench}. Senior experts formulate transformative questions leveraging their extensive experience, while junior and general researchers focus on foundational questions, ensuring a comprehensive benchmark. The annotation process, conducted in controlled laboratory and online settings for consistency, involves sourcing or generating relevant images, designing precise questions, and defining their content and structure using specialized knowledge. Each question undergoes rigorous review by at least five additional expert annotators, whose critical feedback ensures accuracy, clarity, and alignment with the benchmark objectives, safeguarding the integrity and utility of the \textbf{A4Bench}.

\paragraph{\textbf{Question Response}}
Specifically, the example input query to MLLMs can be exemplified as:
\newline
\newline
\textit{\noindent \#User: Which of the following options can describe the affordance of this object as shown in the image? } \newline
\textit{A. This object can be used as a tool for determining directional position \newline  \noindent
B. It can be used as a tool for measuring precise time \newline  \noindent
C. Essential navigational equipment for wilderness exploration \newline  \noindent
D. Used as a tool for drawing accurate circles in drafting \newline  \noindent
E. A small device that operates using Earth's magnetic field}  \newline  
{Answer directly using the letters in the given options.} \newline 
\newline
During evaluation, answer candidates and correct responses are randomized to ensure impartiality. Given that MLLM responses may vary in format, we implement a prompt-human progressive choice evaluation technique to rigorously validate the accuracy of the responses.


\section{Experiment Results}

\begin{figure*}[t]
  \centering
  \centering
    \subfloat[Overall results of \textbf{A4Bench}.]{
        \includegraphics[width=0.55\linewidth]{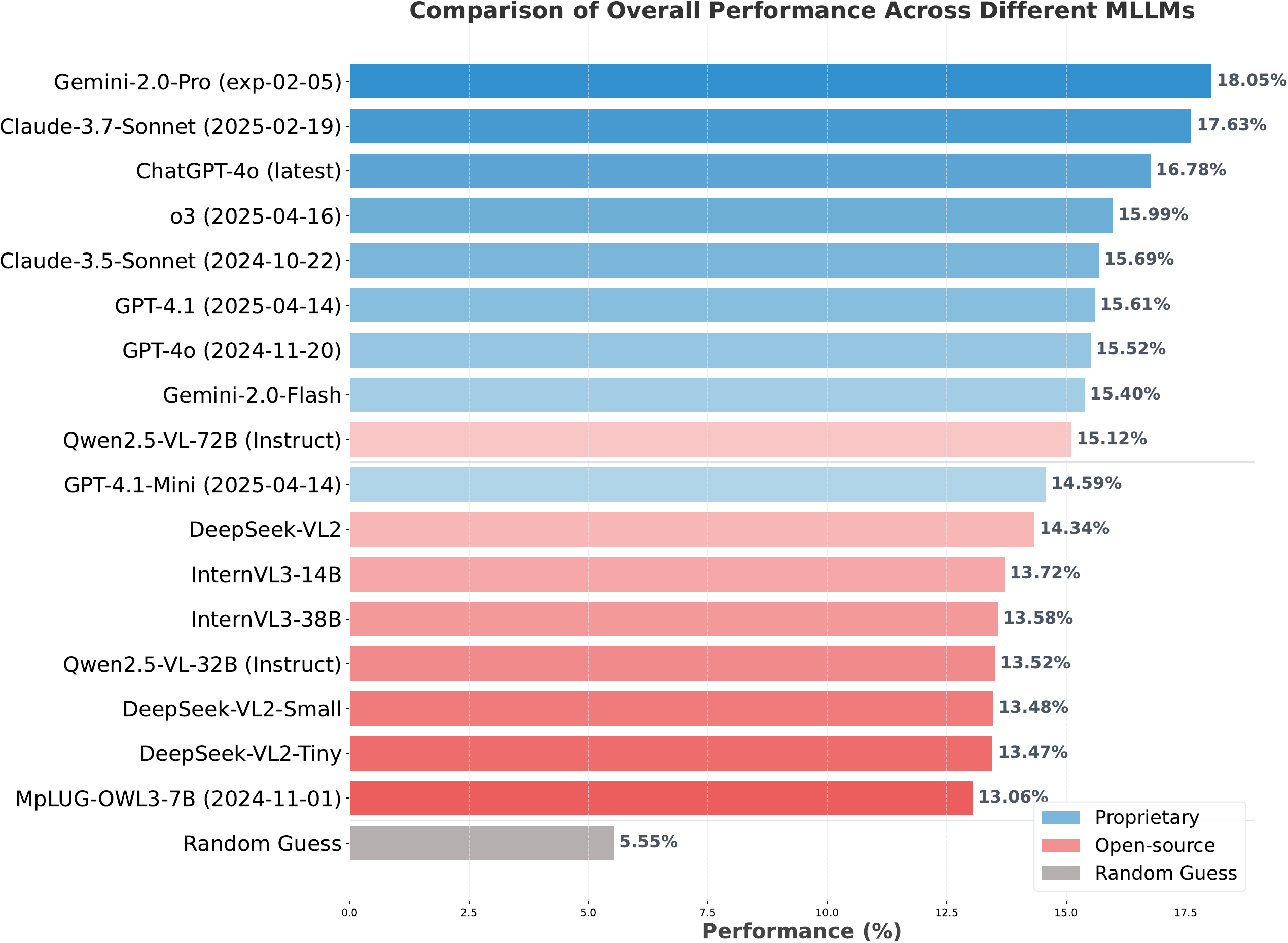}
        \label{fig:performance}
    }
    \hfill  
    \subfloat[Detailed results of \textbf{A4Bench}.]{
        \includegraphics[width=0.43\linewidth]{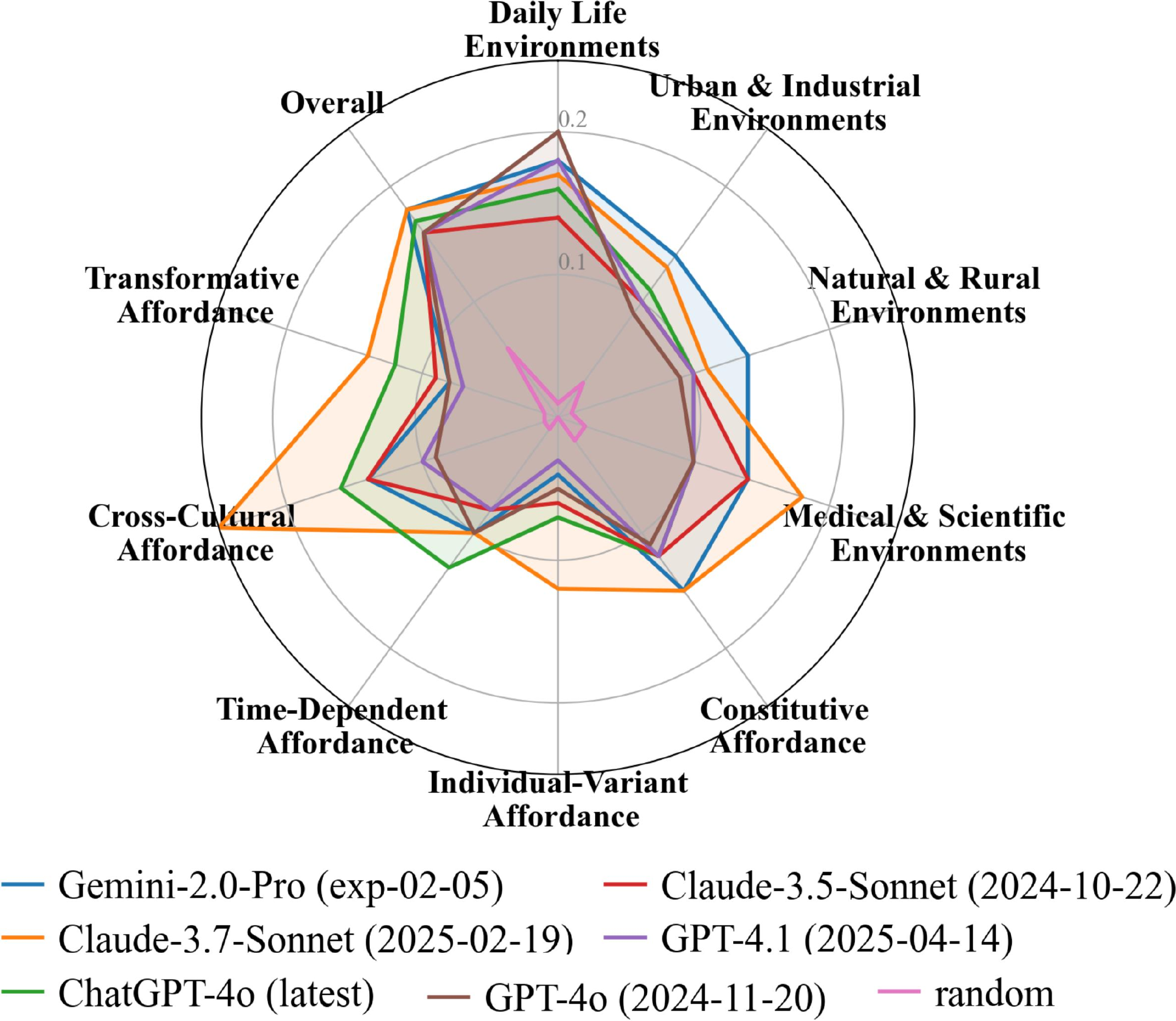}  
        \label{fig:table}
    }
  \caption{ A Quick Look of the A4Bench Outcomes. (a) showcases a comparative analysis of the overall match score between 17 selected MLLMs (both closed-source and open-source), and random guess. (b) displays a radar chart that details the match score performance of the top-7 MLLMs across diverse discplines (Constitutive Affordance) and challenging dimensions (Transformative Affordance excluding Misleading dimensions) of A4Bench.}
  \label{fig:overall}
\end{figure*}

\subsection{Benchmark Candidates}
{In A4Bench, \textbf{9} \textbf{proprietary MLLMs} (\textit{closed-source}) and \textbf{8} \textbf{open-source MLLMs} are choosen for benchmarking. 
The \textbf{Proprietary MLLMs} (\textit{closed-source}) include o3 (2025-04-16) \citep{openai2025o3}, ChatGPT-4o (latest) \citep{openai2024gpt4o}, GPT-4.1 (2025-04-14) \citep{openai2025models}, GPT-4.1-Mini (2025-04-14) \citep{openai2025models}, GPT-4o (2024-11-20) \citep{achiam2023gpt4}, Claude-3.5-Sonnet (2024-10-22) \citep{anthropic2024claude3}, Claude-3.7-Sonnet (2025-02-19) \citep{anthropic2025claude37}, Gemini-2.0-Flash \citep{google2024gemini2}, Gemini-2.0-Pro (exp-02-05) \citep{georgiev2024gemini15}, which are all up-to-date and popular MLLMs. The \textbf{open-source MLLMs} cover DeepSeek-VL2 \citep{deepseekvl2}, DeepSeek-VL2-Small \citep{deepseekvl2}, DeepSeek-VL2-Tiny \citep{deepseekvl2}, InternVL3-14B\citep{internvl3}, Intern-VL3-38B \citep{internvl3}, MpLUG-OWL3-7B (2024-11-01) \citep{mplugowl3}, Qwen2.5-VL-32B (Instruct) \citep{qwen25vl}, Qwen2.5-VL-72B (Instruct) \citep{qwen25vl}, which are all competitive MLLMs. All MLLMs are tested with zero-shot setting. And the instruction prompts (e.g., concept and format prompts in this study) may vary slightly across different MLLMs, tailored to their respective official configurations. }


\begin{table*}[t]\small
    \centering
    \renewcommand\arraystretch{1.1}
    \setlength{\tabcolsep}{5pt}
    \caption{Comprehensive performance comparison of MLLMs and humans on A4Bench across multiple evaluation metrics.}
    \begin{tabular}{l|ccccccc}
    \hline
    \hline
         \textbf{Model} & \textbf{Precision$\uparrow$\quad} & \textbf{Recall$\uparrow$\quad} & \textbf{F1-Score$\uparrow$} & \textbf{Exact Match$\uparrow$} & \textbf{Partial Credit$\uparrow$}  & \textbf{AUC-ROC$\uparrow$} & \textbf{Hamming Loss$\downarrow$} \\ 
    \hline
    \multicolumn{8}{l}{{\textbf{MLLM} \textit{(MLLM)}}} \\ \hdashline
Human (Best) & {93.03\%} & {94.16\%} & {92.90\%} & \textbf{85.34\%} & {88.44\%} & {90.81\%} & {7.19\%} \\
Human (Worst) & {89.73\%} & {91.71\%} & {89.68\%} & {81.25\%} & {84.23\%}  & {88.72\%} & {9.73\%} \\ \hline
\multicolumn{8}{l}{\textbf{Proprietary MLLMs:}} \\ \hdashline
\textsc{o3} & 59.19\% & 35.36\% & 37.40\% & 15.99\% & 36.73\%  & 61.24\% & 42.43\% \\
\textsc{ChatGPT-4o} & 59.17\% & 36.62\% & 38.32\% &  16.78\% & 37.36\% & 62.56\%& 42.25\% \\
\textsc{Claude-3.7-Sonnet} & 61.36\% & 37.74\% & 40.22\% & 17.63\% & 38.09\% & 62.56\% & 40.11\% \\
\textsc{Gemini-2.0-Pro} & 68.27\% & 39.03\% & 41.91\% & 18.05\% & 40.64\%  & 58.52\% & 39.72\% \\
\hline
    \multicolumn{8}{l}{\textbf{Open-source MLLMs:}} \\ \hdashline
\textsc{DeepSeek-VL2} & 57.15\% & 34.57\% & 36.70\% & 14.34\% & 32.57\% & 59.69\% & 44.00\% \\
\textsc{InternVL3-38B} & 56.35\% & 32.95\% & 35.97\% & 13.58\% & 28.95\%  & 56.74\% & 45.30\% \\
\textsc{MpLUG-OWL3-7B} & 55.60\% & 32.91\% & 35.83\% & 13.06\% & 27.91\%  & 55.84\% & 44.38\% \\
\textsc{Qwen2.5-VL-72B } & 59.70\% & 33.42\% & 37.47\% & 15.12\% & 34.06\%  & 60.58\% & 42.69\% \\
 \hline         
\textit{Random guess} & 55.20\% & 57.81\% & 53.16\% &5.55\% & 23.57\%  & 51.63\% & 48.36\% \\
    \hline
    \hline
    \end{tabular}
    \label{tab:main_results}
\end{table*}
\subsection{Human Performance}

To assess human performance on \textbf{A4Bench}, a user study is conducted in a controlled laboratory environment with carefully selected participants. Participants are initially familiarized with the task structure through exposure to representative examples, ensuring comprehension of the evaluation format. They subsequently provide responses to questions presented in \textbf{A4Bench}. To maintain experimental conditions analogous to those of MLLMs, question order is randomized, and participants receive only the concept prompt, format prompt, questions, and answer options, without supplementary information. Both the \textit{best} and \textit{worst} performance outcomes are documented for comparative analysis. This rigorous methodology ensures reliable and results. Statistical measures are applied to evaluate response consistency.

\subsection{Findings of \textbf{A4Bench}}

\paragraph{\textbf{Proprietary MLLMs vs. Open-Source MLLMs vs. Human}} A concise overview of the \textbf{A4Bench} results is provided in Figure \ref{fig:overall}, which offering several key insights: 1) Proprietary MLLMs consistently outperform their open-source counterparts, with models like Gemini-2.0-Pro achieving an overall score of 18.05\%, surpassing the best open-source model, Qwen2.5-VL-72B, at 15.12\%. This gap underscores the advantage proprietary models have, likely due to access to more extensive and diverse training datasets. Although the performance margin is not always vast and the first open-source models demonstrate competitiveness, proprietary systems currently maintain an advantage in overall affordance understanding as gauged by this benchmark. 2) All evaluated MLLMs show significantly limited capabilities in affordance perception, which is substantially below human-level performance. The top-performing MLLM, Gemini-2.0-Pro, lags behind the human best by 14.62\%, and even the human worst at 81.25\% surpasses the best MLLMs by 3.39\%. This stark disparity highlights that current MLLMs, despite their advancements, struggle to replicate human-level understanding of affordance-related tasks, necessitating further development in this area. 3) All models outperform the random guess baseline, indicating some level of learned capacity, although insufficient for complex affordance percception and understanding.

As shown in Table~\ref{tab:main_results}, we further report comprehensive performance metrics beyond exact match accuracy, including Precision~\cite{STREINER2006327}, Recall~\cite{10.1145/1143844.1143874}, F1-Score, Partial Credit, AUC-ROC, and Hamming Loss, to provide a more nuanced evaluation. The results reveal several key findings: 1) Human performance substantially surpasses all MLLMs across every metric. 2) Proprietary MLLMs consistently outperform open-source counterparts, with Gemini-2.0-Pro leading among models. 3) Despite moderate Precision values, low Recall severely constrains the F1-Score, indicating difficulty in consistently identifying affordance. 4) Random guess achieves surprisingly high baseline Precision and Recall, but its extremely low Exact Match demonstrates poor practical utility. In general, these findings highlight the urgent need to advance the capacity of MLLMs to perceive context-sensitive and dynamic affordance.

\paragraph{\textbf{Findings of Constitutive Affordance}} 
In the constitutive affordance category, which evaluates the ability to perceive affordance in static contexts, performance varies significantly across disciplines. 1) Proprietary models demonstrate a clear edge, with ChatGPT-4o and GPT-4-1 scoring 16.41\% and 18.18\% in the \textbf{Daily discipline}, respectively, indicating robust performance in understanding affordance in everyday scenarios. Similarly, in the \textbf{Natural discipline}, both models achieve 10.45\%, reflecting an ability to handle natural environments effectively. However, their performance dips in more specialized disciplines, such as \textbf{Medical} (9.63\% for ChatGPT-4o) and \textbf{Urban} (10.98\% for GPT-4.1), suggesting limitations in domain-specific contextual understanding. 2) In contrast, open-source models such as DeepSeek-VL2 and InternVL3-38B demonstrate consistent yet comparatively lower performance across these disciplines, with scores ranging from 8.65\% to 9.63\% and 8.60\% to 9.30\%, respectively. This uniformity suggests a balanced but less specialized capability relative to proprietary models. 3) All MLLMs demonstrate suboptimal performance relative to human capabilities, exposing a substantial gap in interpreting constitutive affordance with the depth, nuance, and precision inherent in human understanding.


\paragraph{\textbf{Findings of Transformative Affordance}} 
In the transformative affordance category category evaluates the understanding of dynamic and contextually shifting affordance. The overall score for this category is calculated as a weighted average of the Time-Independent, Cross-Culture, and Individual dimensions. The exclusion of the Misleading dimension arises because its random guess rate is notably high, necessitating the subtraction of this baseline to accurately reflect the model true performance in that dimension. Performance disparities across dimensions emerge, as depicted in Table \ref{tab:part1}. Specifically, 1) proprietary models like Gemini-2.0-Pro lead with a \textbf{Time-Dependent} score of 15.2\% and a \textbf{Culture score} of 13.56\%, demonstrating some proficiency in handling temporal transformations and cultural nuances. However, they struggle significantly in the \textbf{Misleading} dimension (67.38\%) and the \textbf{Individual} dimension (4.43\%), indicating challenges in detecting deceptive affordance and interpreting individual-specific contexts. Similarly, Claude-3.5-Sonnet achieves a balanced performance with 14.86\% overall but shows weaknesses in \textbf{Individual} at 6.40\%. 2) Open-source models, such as Qwen2.5-VL-72B, display more uniform but lower scores across these dimensions, ranging from 8.10\% in \textbf{Misleading} to 9.32\% in \textbf{Culture}, with a particularly poor \textbf{Individual} score of 5.42\%. This suggests that open-source models lack the depth required for nuanced transformative affordance perception. 3) In comparison, human performance remains consistently high, ranging from 83.67\% (Time-Dependent) to 84.75\% (Culture), highlighting a significant gap. The underperformance of MLLMs in transformative affordance tasks, especially in dimensions requiring complex reasoning like \textbf{Misleading} and \textbf{Individual}, highlights the imperative for enhanced training strategies to improve their capacity to address dynamic and subjective affordance scenarios.


\section{Conclusion}
This paper introduces \textbf{A4Bench}, a novel and comprehensive benchmark featuring 2,000 multimodal question-answer pairs, specifically designed to systematically assess the affordance perception capabilities of MLLMs across two critical dimensions: constitutive affordance and transformative affordance. Our comprehensive evaluation of 17 MLLMs compared to human performance reveals substantial limitations in current model capabilities. All models significantly underperform relative to human understanding, particularly in the nuanced domain of transformative affordance. These findings reveal critical challenges for MLLMs in grasping contextual and dynamic affordances. A4Bench addresses this gap by providing a rigorous diagnostic framework to identify weaknesses in perception and reasoning. Furthermore, it offers a standardized benchmark to foster fair comparison, drive innovation, and guide the development of safer, more context-aware multimodal AI systems.

\bibliography{aaai2026}

\end{document}